# Decision-Making in Reinforcement Learning

*Arsh Javed Rehman, **Dr. Pradeep Tomar
*Student, Dept. of Computer Science & Engineering, Gautam Buddha University
**Asst. Professor, Dept. of Computer Science & Engineering, Gautam Buddha University

*Abstract*

*In this research work, probabilistic decision-making approaches are studied, e.g. Bayesian and Boltzmann strategies, along with various deterministic exploration strategies, e.g. greedy, ϵ-Greedy and random approaches. In this research work, a comparative study has been done between probabilistic and deterministic decision-making approaches, the experiments are performed in OpenAI's gym environment, solving Cart Pole problem.*

*This research work discusses about the Bayesian approach to decision-making in deep reinforcement learning, and about dropout, how it can reduce the computational cost. All the exploration approaches are compared. It also discusses about the importance of exploration in deep reinforcement learning, and how improving exploration strategies may help in science and technology.*

*This research work shows how probabilistic decision-making approaches are better in the long run as compared to the deterministic approaches. When there is uncertainty, Bayesian dropout approach proved to be better than all other approaches in this research work.*

***Keywords:*** *Reinforcement Learning, Deep Q Network, Bayesian Approach, Exploration and Exploitation Trade-Off, Deep Reinforcement Learning*

1. **Introduction**

Reinforcement learning is referred to as learning what to do, in order to maximize rewards and map the situations to actions. The learner or agent does not know which actions to perform, instead discover actions which yield the maximum rewards by trying them. The actions performed may or may not fetch the immediate reward but may yield reward at a longer time. The important features of RL are trial-and-error search and delayed reward. [2]

2. **Exploration and Exploitation**

To make a decision we have to make a choice between Exploitation and Exploitation. *Exploitation is making the best decision from the given information* and *Exploration is to gather more information*. Let's take an example of restaurant selection, which involves a decision, we can choose Exploitation and go to our most loved restaurant or can-do Exploration that is try at a new restaurant.[3]

Exploration is needed for an agent, in order to learn all available states present in an environment and how to deal with it optimally, agent must know as many states as possible. In a RL problem, agent has limited access via its actions on the environment unlike other traditional supervised learning approaches. As an outcome, there occurs a problem that is, for an agent to learn optimum policy it requires the right amount of experiences, but for obtaining those experiences requires a good policy.[1]

As a result, arises a challenge in reinforcement learning that is the trade-off between exploration and exploitation. A RL agent must perform actions from its experience that are actions taken from the past experience which had proved effective for obtaining rewards. But in order to explore those actions, agent must try actions that it has not performed in the past. To obtain rewards the agent will exploit that is actions from its knowledge and experience of the past, but for better rewards in the future and better action-selection, the agent has to explore. Different actions must be tried by the agent and steadily favour those actions that seems to be best. In case the task be Stochastic, every action must be repeated and tried several times to gain experience and valid approximation of its likely reward. The dilemma of exploration and exploitation has been studied for decades intensively by the and mathematicians and researchers and still it remains unresolved. As of now, the exploration and exploitation issue does not arise in other machine learning categories i.e. unsupervised and supervised learning.

Exploitation is preferred in order to maximize the likely rewards but exploration gives us the opportunity to acquire the greater reward in the long run. Let us suppose for example, during exploration, in the short run reward is less, but in the long run reward is higher because after the agent has explored better actions, the agent can exploit for better reward. [5]

### 3. Dropout as a Bayesian Approximation Approach

This approach was given by Yarin Gal[4]. In this approach the uncertainty of actions is exploited by the agent. This approach uses the Bayesian Neural Networks or BNNs, it is a type of neural network that act probabilistically rather than deterministically. There is no set of fixed weights rather a probability distribution is used in assigning weights. The probability distribution of weights helps in distribution of actions in a reinforcement learning environment. The uncertainty of taking an action by an agent is given the variance of probability distribution.

This approach comes with a prohibitive cost of computation and maintaining a distribution over all weights is impractical. In order to stimulate the neural network, we use dropout. In the training process the activation nodes are set randomly to zero, this technique is called Dropout. Uncertainty about every action are obtained by repeating the process of sampling.

Bayesian models provide a mathematical framework to find out about the uncertainty of a model. Dropout in Bayesian helps in training and reducing computational cost. This approach helps in representing uncertainty without losing test accuracy.

### 4. Other Exploration Approaches

Other frequently used exploration approaches in Reinforcement Learning are:

*4.1 Greedy Approach*

The goal of all RL algorithms to is attempt and maximize reward in the long run. An inexperience or simple approach is to choose action that is optimal at any point of time which gives the agent the highest reward. This approach is referred as a greedy method. This approach does not use any exploration rather exploit its knowledge. The action which is best at that time is taken by the agent providing it with highest reward. This method does not explore other long-term rewards. The shortcoming of this approach is that it converges to a suboptimal solution. For example, in a two-armed bandit [5], let us suppose the reward is different for both the arms, one arm rewards 1 point and other arm rewards 2 points, suppose the agent due to its parameters chooses 1 point arm then under this approach the agent will never explore, and will never choose the other arm.

*4.2 Random Approach*

Random Approach is just the opposite of greedy method. It simply selects the actions randomly. Does not care about the reward. This approach is ideal when a random policy is required. This method is used for sampling and getting experience by taking random actions. In DQN experience buffer can be filled using this approach.

*4.3 Epsilon($\epsilon$)-Greedy Approach*

This approach is popularly used for exploration. It is a combination of random and greedy approach. The agent chooses action which provides the highest reward and sometimes chooses action randomly, so it both explores and exploits the action. In this method the agent acts optimal and agent may take new actions for better experience. The Epsilon ($\epsilon$) in this approach can be changed accordingly, changing epsilon changes the probability of choosing random action. This approach has been used in several recent RL algorithms including DQN [1] because of its accuracy and simplicity, it has now become a de facto method. The value of epsilon is initialized for higher probability to promote exploration during training period. The value starts degrading slowly and annealed down to a small value usually to 0.1, by this point agent has enough experience and can exploit what it has learnt. The shortcoming of this approach is that it is still far from optimal solution.

*4.4 Boltzmann Approach*

For a better exploration, preferably the agent must exploit all the states and actions and exploit Q-values generated by the network. This is done in Boltzmann Approach. In this approach action is chosen according to its probability rather than choosing an optimal action all the time, or choosing action randomly. To choose an action we use a SoftMax function that estimates the probability of each action. The most likely action is taken by the agent using this approach. This approach considers all the actions information. It is better than epsilon greedy approach, as it considers all the actions according to their relative weight while epsilon greedy considers all the actions equally. For example, an agent can perform 4 actions, then all the 4 actions will be considered according to their relative weight while in epsilon greedy approach even the 3 non-optimal solution will be considered equally by the agent. This approach helps the agent ignore sub-optimal solution and focus on optimal solution. During training a temperature parameter is used to control the SoftMax distribution, so that all the actions to be considered equally at the start of training. The shortcoming of this approach is that SoftMax gives the value to an action, if action 1 is 0.8 and action 2 is 0.1, the agent thinks that action 1 is 80% optimal while action 2 is only 10% optimal. This is not the case in reality. It is not the best approach for exploration.[6]

5. **Results**

All the above exploration approaches are compared in OpenAI's gym environment - CartPole and implemented using Deep Q-Networks. In all the approaches Boltzmann and Bayesian Dropout Approximation gives better results in my experiment. The code and implementation can be found on [https://github.com/arshjrehman/RL.git], better results can be obtained by changing the hyperparameters.

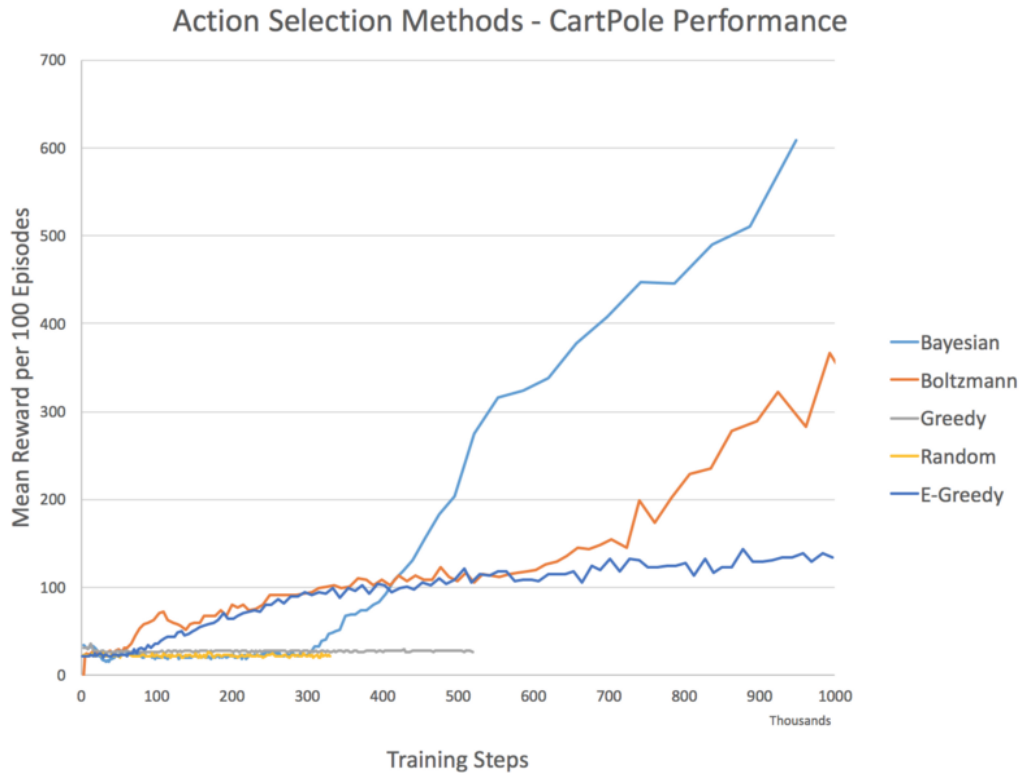

## 6. Conclusion

In reinforcement learning, balancing exploration and exploitation is a major challenge, a RL agent must perform actions from its experience that are actions taken from its past experience which had proved optimal for obtaining rewards. In this paper we discussed various exploration approaches. Bayesian Dropout is compared with various other exploration approaches. Bayesian Dropout proved to be better than all other exploration approaches. Better exploration will help the agent make better decision and deal optimally in the environment.